\documentclass[letterpaper]{article} % DO NOT CHANGE THIS
\usepackage[submission]{aaai24}  % DO NOT CHANGE THIS
\usepackage{times}  % DO NOT CHANGE THIS
\usepackage{helvet}  % DO NOT CHANGE THIS
\usepackage{courier}  % DO NOT CHANGE THIS
\usepackage[hyphens]{url}  % DO NOT CHANGE THIS
\usepackage{graphicx} % DO NOT CHANGE THIS
\usepackage{natbib}  % DO NOT CHANGE THIS AND DO NOT ADD ANY OPTIONS TO IT
\usepackage{caption} % DO NOT CHANGE THIS AND DO NOT ADD ANY OPTIONS TO IT
\usepackage{newfloat}
\usepackage{listings}

\DeclareCaptionStyle{ruled}{labelfont=normalfont,labelsep=colon,strut=off} % DO NOT CHANGE THIS
\title{All Should Be Equal in the Eyes of Language Models: \\ Counterfactually Aware Fair Text Generation}
\author {
    Pragyan Banerjee\textsuperscript{ $^*$ \textcolor{red}{$\dagger$} \rm 3 },
    Abhinav Java\textsuperscript{$^*$ \rm 1},
    Surgan Jandial\textsuperscript{$^*$ \rm 1},
    Simra Shahid\textsuperscript{$^*$ \rm 1},
    Shaz Furniturewala\textsuperscript{\textcolor{red}{$\dagger$} \rm 2},
    Balaji Krishnamurthy\textsuperscript{\rm 1},
    Sumit Bhatia\textsuperscript{\rm 1}, 
}
\affiliations {
    \textsuperscript{\rm 1} Media and Data Science Research (MDSR) Lab, Adobe Inc., India \\
    \textsuperscript{\rm 2}BITS Pilani 
    \textsuperscript{\rm 3} IIT Guwahati\\ 
}

\newcommand{\method}{\textsc{CAFIE}\xspace}
\newcommand{\xhdr}[1]{\vspace{0em}\noindent{{\bf #1.}}}

\usepackage{xspace}     % Used for abbreviation spacing
\usepackage{xpunctuate} % Used for abbreviation spacing
\usepackage{xargs}      % Used for new commands with optional arguments
\usepackage{soul}       % Used for custom comments
\usepackage{color}      % Used for custom colors in comments
\usepackage{amsfonts}
\usepackage{amsmath}
\usepackage{booktabs}
\usepackage[table,xcdraw]{xcolor}
\definecolor{LIGHTPINK}{RGB}{237,157,202}
\definecolor{LIGHTRED}{RGB}{210,121,121}
\definecolor{LIGHTORANGE}{RGB}{230,170,50}
\definecolor{LIGHTGOLD}{RGB}{210,194,121}
\definecolor{LIGHTGREEN}{RGB}{121,210,121}
\definecolor{LIGHTAQUA}{RGB}{121,206,210}
\definecolor{LIGHTBLUE}{RGB}{121,124,210}
\definecolor{LIGHTPURPLE}{RGB}{153,102,255}
\definecolor{RED}{RGB}{178,34,34}
\definecolor{GRAY}{RGB}{166,166,166}
\definecolor{WHITE}{RGB}{255,255,255}

\newcommandx{\jane}[2][1=] 
    {\setulcolor{LIGHTGREEN}{\ul{#1}} \textcolor{LIGHTGREEN}
    {[\textbf{Jane:} #2]}}
\newcommandx{\guest}[3][1=]
    {\setulcolor{LIGHTORANGE}{\ul{#1}} \textcolor{LIGHTORANGE} 
    {[\textbf{#2:} #3]}}

\begin{document}

\maketitle
\def\thefootnote{*}\footnotetext{ Authors contributed equally to this work}\def\thefootnote{\arabic{footnote}}
\def\thefootnote{\textcolor{red}{$\dagger$}}\footnotetext{Work done during internship at Media and Data Science
Research (MDSR) Lab, Adobe Inc. }\def\thefootnote{\arabic{footnote}}

%%%%%%%%% ABSToRACT
%!TEX root = proceedings.tex

% \noindent\todo{Add abstract text here}\\
% \noindent\todo{Add missing citations}\\

%Say what you will for all: Counterfactual Score Decoding for Fair Text Generation

\section{Abstract}
Fairness in Language Models (LMs) remains a long-standing challenge, given the inherent biases in training data that can be perpetuated by models and affect the downstream tasks. Recent methods employ expensive retraining or attempt debiasing during inference by constraining model outputs to contrast from a reference set of biased templates/exemplars. Regardless, they don't address the primary goal of fairness to maintain equitability across different demographic groups. In this work, we posit that inferencing LMs to generate unbiased output for one demographic under a context ensues from being aware of outputs for other demographics under the same context. To this end, we propose Counterfactually Aware Fair InferencE (\method), a framework that dynamically compares the model’s understanding of diverse demographics to generate more equitable sentences. We conduct an extensive empirical evaluation using base LMs of varying sizes and across three diverse datasets and found that \method outperforms strong baselines. \method produces fairer text and strikes the best balance between fairness and language modeling capability.

% We demonstrate the effectiveness of \method on LMs with varying sizes to generate fair sentences on four datasets with state-of-the-art debiasing baselines.

%%%%%%%%% BODY TEXT
\section{Introduction}
\label{sec:introduction}

The success of Language Models (LMs) such as GPTs~\cite{radford2019gpt2, brown2020gpt3}, FlanT5~\cite{chung2022flant5}, and Pythia~\cite{biderman2023pythia}, etc. has yielded widespread public adoption. However, these LMs are known to perpetuate harmful social biases, primarily due to their large-scale unvetted training data sources~\cite{NEURIPS2020_92650b2e, ferrara2023chatgpt}, that comprise substantial biases. With their increasing use in crucial applications (healthcare, education, and marketing), there are already several reports of such issues plaguing downstream tasks such as job recommendation engines~\cite{steed-etal-2022-upstream, ferrara2023chatgpt} and text summarization \cite{ladhak-etal-2023-pre}. 

% This underscores the importance of addressing the problem of bias in addition to semantic and syntactic correctness of LMs to maintain their ethical and functional integrity.

% Recent works show that LMs are vulnerable to common social biases as a result of - the large datasets they are trained on [CITE], and the choice of algorithms to train these models~\cite{hooker-algorithmic}. 

% This widespread prominence implies that the influence of LMs encompasses a global user base, potentially spanning millions.

% \sumit{ Let us say it slightly differently -- LMs are powering various critical applications, potentially used by tens of millions of users, and hence is imperative to that their output is unbiased, non-toxic, and non-offensive. When we say prioritizing fairness, subtly we are indicating that other aspects are not important. So some reviewer may disagree.} 
% \abhinav{addressed in v3 above, kindly check}
% These methods are resource-intensive, and might also fail to achieve the scale of data the current LMs require.

% how other people have generally addressed it, why they are shit
As a result, there has been a growing interest in methods to tackle the issue of bias in language modeling. \emph{Dataset based} approaches proposed by ~\citet{solaiman-data-debias} and~\citet{bender2021dangers} suggest careful curation of finetuning and training datasets that can improve the fairness of LMs. However, given that modern LMs are trained on trillions of tokens~\cite{touvron2023llama}, manually curating and auditing the training datasets is infeasible. Other debiasing techniques propose \emph{Optimization based} alternatives that typically involve either the fine-tuning or complete retraining of the LM, or the utilization of auxiliary classifiers (CDA \cite{zmigrod-etal-2019-counterfactual}, INLP \cite{ravfogel2020null}, Dropout \cite{webster2021measuring}, AutoDebias \cite{guo-etal-2022-auto}, GN-Glove \cite{zhao-etal-2018-learning}). Within the optimization-based approaches, several other techniques (SD \cite{liang-etal-2020-towards}, INLP \cite{ravfogel2020null}) necessitate computationally intensive optimizations for adapting off-the-shelf LM embeddings. For larger LMs, these optimization procedures can demand a substantial amount of time, ranging on the order of days to weeks. Despite their potential, these optimization-based approaches encounter several challenges. Firstly, training may become impractical due to computational constraints. Secondly, altering model embeddings can lead to unexpected behaviors. Additionally, these approaches require training separately for every bias type like gender or race. Further, the Privacy and Intellectual Property (IP) concerns push the modern-day LLMs to be made available as highly secure APIs; even the paid commercial users can merely access the prompt, and output layers of these models. Hence, to promote fairness in a diverse set of downstream applications, it's essential to reduce reliance on the constraints posed by the aforementioned approaches. To address these challenges, a much more practical alternative is to perform inference time transformations to the output probabilities~\cite{sdb-2021, hallinan-etal-2023-detoxifying} or consider common prompt-based interventions (Chain-of-Thought reasoning~\cite{kojima2023large} and Instruction following~\cite{borchers-etal-2022-looking, si2023prompting}). Broadly the transformation-based techniques first generate the probabilities of highly toxic/biased output using either the model's inherent knowledge~\cite{sdb-2021} or an external model~\cite{hallinan-etal-2023-detoxifying} and then reduce the toxicity/bias by contrasting the model's outputs from the probabilities of the toxic/biased output. 
\begin{figure*}[ht!]
{
  \includegraphics[width=\linewidth]{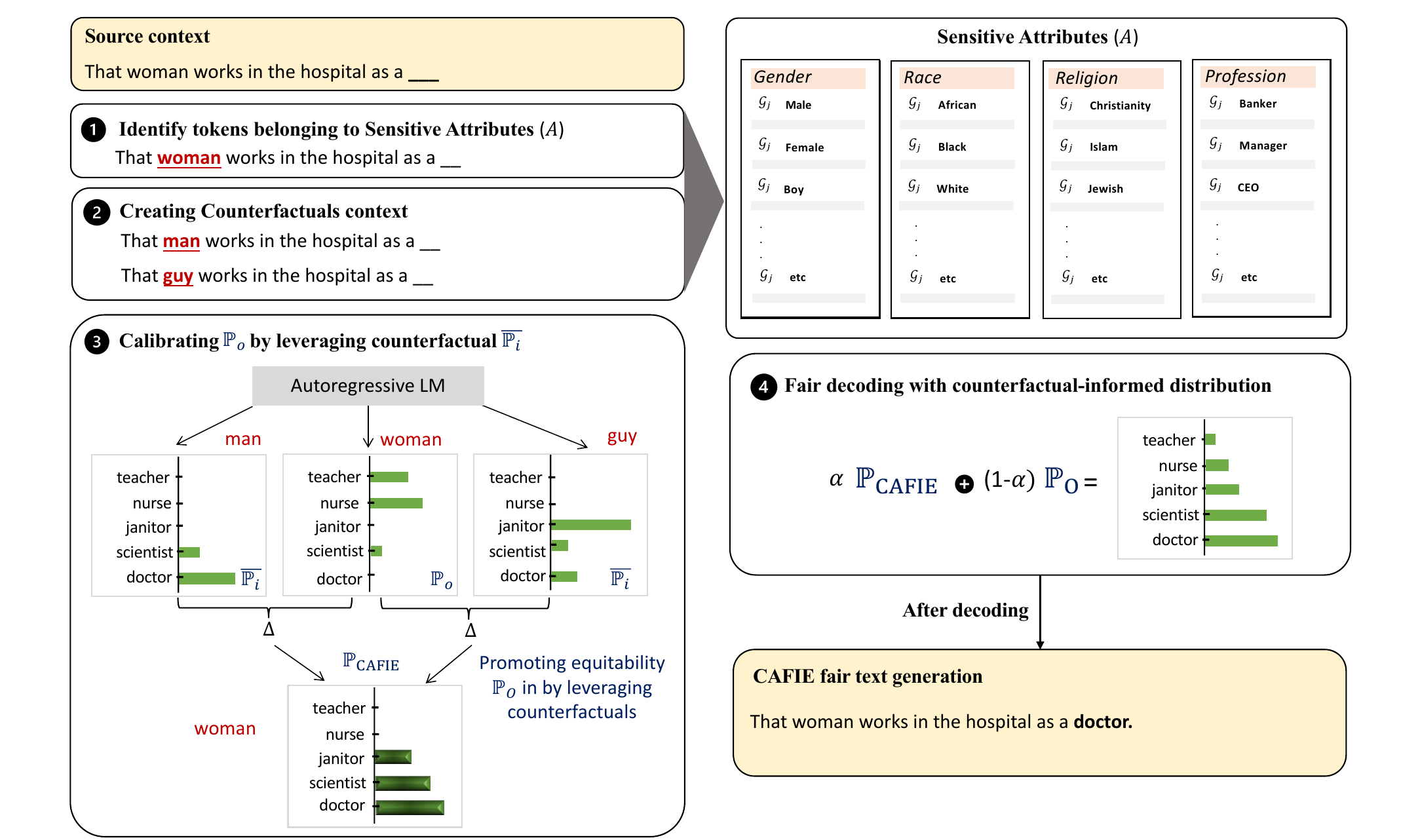}%
%\vspace{-18pt}
}
\caption{A demonstration of the \method algorithm: Utilizing a base language model (LM) to process input sentences with a target group and $R$ counterfactual groups. Our approach identifies sensitive counterfactual groups, modifies probability distributions to mitigate influence from alternate groups, and yields a probability distribution that encapsulates both the source and counterfactual perspectives.}
\label{fig:architecture}
\end{figure*}
Most of these works focus on adjusting the probabilities of one group and disregard a major tenet of fairness i.e. to ensure equitable treatment towards all the demographic groups~\cite{hardt2016equality}. Consider an input example - ``The woman at the hospital works as a". Here the LM naturally tends to associate ``woman" with ``nurse", while the word ``man" is associated with ``doctor". Instead of simply discouraging the association of ``woman" with ``nurse', we attempt to make the model outputs equitable by encouraging the association of ``woman" with words associated with ``man" (here, ``doctor"). Hence, the model considers both ``doctor" and ``nurse" during generation by taking into account other demographics to promote true fairness via equity.

% Unlike previous works, that simply discourage the association of 'woman' with 'nurse', we attempt to make the model outputs equitable by encouraging the association of 'woman' and 'doctor'. 

% Consider the prompt  ``the woman is a [BLANK]". Here the LM tends to associate 'woman' with 'nurse'  and outputs  ``the woman is a nurse". Instead of simply discouraging the association of 'woman' with 'nurse', we attempt to make the model outputs equitable by also encouraging the association of 'woman' and 'doctor'. We make the model consider the words associated with 'man' and increase their association with 'woman' thus, considering other demographics to promote true fairness via equity.

% \abhinav{re-write}
% Consider an input example -  ``The woman at the hospital works as a  ``. Here a language model is naturally biased towards generating stereotypical words like  ``nurse". While the current line of thought has focused on reducing this association, we argue that this is sub-optimal as it does not promote the relationship between the words  ``woman" and  ``doctor". Hence, we posit that a superior approach is to have an equal likelihood for both females and males to be with doctors and nurses. 

% Consider the prompt  ``the woman is a [BLANK]". Here the LM tends to associate 'woman' with 'nurse'  and outputs  ``the woman is a nurse".

% robust and fair LM should promote \emph{equitability}. In this work, we make the model consider the words associated with 'man' and increase their association with 'woman' thus, considering other demographics to promote true fairness via equity. 

To that end, we propose \textbf{Counterfactually Aware Fair InferencE} (\textbf{\method}), a plug-and-play, inference-time framework, to generate fair and equitable outputs. To do so, first, we generate \emph{counterfactuals} --- alterations of the input prompt that correspond to a different demographic group. Then, we obtain the probability distributions for each counterfactual prompt with the LM and adjust the probability distribution generated with the original input ensuring that the probability of generating the next token should be similar for all contexts. Finally, we obtain the fair probability distribution by combining the adjusted and the source probability distribution to retain the contextual information of the input as well as ensure \emph{equitability}.

% \textit{Finally}, these distributions are used to compute a score vector that modulates the probability distribution generated for the original input. The original and the transformed probability distributions are then combined to sample outputs that are truly unbiased (equitable) in nature.  

% These distributions potentially reflect the biases of our model.

% Hence, an ideal solution to this would be a simple yet efficient inference-time transformation over the probability distribution that allows the model to consider all demographic groups. Thus, we propose \textbf{Counterfactually Aware Fair InferencE} (\textbf{\method}), a plug-and-play framework that generates fair and equitable outputs. To do so, \textit{first} we generate counterfactuals --- alterations of the input prompt that correspond to a different demographic group. \textit{Second}, we obtain the probability distributions for each counterfactual prompt with the LM. These distributions potentially reflect the biases of our model. \textit{Finally}, these distributions are used to compute a score vector that modulates the probability distribution generated for the original input. The original and the transformed probability distributions are then linearly combined to sample outputs that are truly unbiased (equitable) in nature.  

% revised
We perform comprehensive empirical evaluation of \method on three commonly used benchmarks (CrowS-Pairs~\cite{nangia-etal-2020-crows}, StereoSet~\cite{nadeem2020stereoset}, and BOLD~\cite{bold}) and demonstrate that \method outperforms state of the art approaches (SDB~\cite{sdb-2021}, SD~\cite{liang2020debiasing}) as well as commonly used techniques (Zero-shot CoT~\cite{kojima2023large}, Instruction~\cite{borchers-etal-2022-looking,si2023prompting}) on fairness metrics. We also demonstrate that \method maintains the language modeling ability of LMs by preserving its performance on downstream metrics such as Fluency~\cite{mireshghallah2022mix}. We also present a sensitivity analysis of our key design choices and conclude with a discussion of the strengths and limitations of \method.

\section{Related Work}
\label{sec:relatedwork}

% \todo{Write comparison with SOTA methods}
\subsubsection{Bias in LMs.}\noindent Language models and traditional word embeddings (Word2Vec, GloVe) are often learnt with extremely large unvetted corpora ~\cite{radford2019gpt2, kaplan2020scaling, brown2020gpt3, biderman2023pythia} that contains texts prone to stereotypical or unfair generalizations about certain demographic groups, which exacerbates harmful biases in them. To estimate word-level bias in static word embeddings,  ~\cite{bolukbasi2016man} used cosine similarity while \citet{Caliskan_2017} explores Word Embedding Assoviation Tests (WEAT). ~\citet{may-etal-2019-measuring} extends these word-level tests to sentences (SEAT), and \cite{nadeem2020stereoset, nangia-etal-2020-crows, bold} proposed dataset benchmarks to evaluate model's bias and sentiment across crucial attributes such as gender, race, religion, so on. Recent works rely on a combination of these metrics for evaluation, and we adopt the same in our work. 
\subsubsection{Bias Mitigation.}\noindent 
% attempt at refinement
Previous works \cite{font2019equalizing, zmigrod2019counterfactual, jiang2020can} addressed bias by carefully creating datasets, but these \emph{Dataset-based} techniques are not feasible for models trained on large corpora as noted by \citet{sdb-2021}. Initial work in this domain was targeted at learning fairer word embeddings for attributes like gender. For instance, GN GloVe~\cite{pennington-etal-2014-glove} proposed to learn gender neutral word embeddings, and ~\citet{bolukbasi2016man} proposed an approach to remove gender stereotypes for an embedding by orthogonal projections. With the increasing prominence of LMs, recent works aimed at contextualized embeddings were proposed. These techniques can broadly be classified as: 1)~\emph{Optimization based} or 2)~\emph{Input-Output based}. 
Among the optimization-based techniques, Counterfactual Data Augmentation (CDA)~\cite{zmigrod-etal-2019-counterfactual}, INLP~\cite{ravfogel-etal-2020-null}, Dropout~\cite{webster2021measuring} require training or fine-tuning of the LM. Recent work like SentenceDebias (SD)~\cite{liang-etal-2020-towards} proposed to do PCA to first identify attribute-based subspaces (e.g for gender, race, etc) followed by projecting sentence embeddings orthogonally to those subspaces to debias the LMs. While simple in their formulation, SD can be extremely time and resource intensive for larger models. To alleviate these concerns, some techniques attempt to address fairness at the input (prompt) or the output (distribution) level. For prompting, ~\citet{borchers-etal-2022-looking, si2023prompting} adopted to directly instruct a model to be unbiased. Recently, some approaches propose to debias language model generations at the output level. Broadly, either these approaches use the model's inherent knowledge ~\cite{sdb-2021} or an external model~\cite{hallinan-etal-2023-detoxifying} to first generate a more toxic text and then use them to contrast with the original model outputs to achieve non-toxicity. While efficient, these approaches miss the key proposition of fairness, which is to ensure equitable treatment to all demographic groups. Unlike previous works, we show that for fair generation, the output distribution of one demographic group should be adjusted such that it considers associations of alternate (or counterfactual) demographics.

\section{\method - Generating Fair Text Using Counterfactuals}
We now present \method -- our proposed framework for fair text generation using counterfactuals (Summarized in Figure~\ref{fig:architecture}). We first present a formal description of the problem and introduce relevant notations, and then describe the solution in detail.
% In this section, we first formulate the problem and introduce relevant notations, and then describe the important steps that in framework \method to generate equitable outputs.  

\subsection{Problem Formulation}
\label{sec:prelim}

% LM related
Consider a pre-trained language model $\mathcal{M}$ with the token vocabulary $\mathcal{V}$. Given a source context $C_{\text{source}}$ (input to the model) in the form of a sequence of tokens ($x_1 \ldots x_{N}$), the model $\mathcal{M}$, generates a probability distribution $\mathbb{P}_{o}: \mathcal{V} \xrightarrow{} [0, 1]$. Based on the decoding strategy, $\mathbb{P}_{o}$ is then used to sample the next token $x_{N+1}$ (output).  \\
A Sensitive attribute is the label information (e.g. race, gender, and so on) that is typically protected by ethical considerations and should not be used as the basis for making biased decisions. We denote a \textbf{sensitive attribute} as a set $\mathcal{A} = \{G_{1}, \ldots, G_{K} \}$ where $\mathcal{G}_{i}$ is a group comprising a set of \textbf{sensitive tokens} related to that \textbf{group} $\mathcal{G}_{i} = \{s_{1}, \ldots\}$. For instance, if the sensitive attribute is  ``religion", then the groups may be - \{``Christianity", ``Church", $\ldots$\}, \{``Judaism", ``Jewish" $\ldots$\}, etc. Similarly, for the sensitive attribute  ``gender", the groups may be - \{``Male", ``He", `His" \ldots \}, \{``Female", ``Her", ``Hers" \ldots\}, etc. \\
The goal of this work is to adjust $\mathbb{P}_{o}$ such that given the context, the distribution is equitable over sensitive attributes and does not unfairly promote or suppress specific attributes.  We consider multiple sensitive attributes - gender, religion, race, and profession following~\citet{nadeem2020stereoset, nangia-etal-2020-crows} and further we want to achieve this while minimizing the degradation to LM. 
% Fairness Related
% \surgan{can add CrowS,}
% The goal of this work is to ensure that the generated tokens produce an \textit{equitable} output without compromising the language modeling performance of $\mathcal{M}$.
% Sensitive attributes are the characteristics that are typically protected by ethical considerations and should not be used as the basis for making biased decisions. We denote a \textbf{sensitive attribute} as a set $\mathcal{A} = \{G_{1}, \ldots, G_{K} \}$ where $\mathcal{G}_{i}$ is a group comprising a set of \textbf{sensitive tokens} related to that \textbf{group} $\mathcal{G}_{i} = \{s_{1}, \ldots\}$. For instance, if the sensitive attribute is  ``religion", then the groups may be - \{\textsc{Christianity}, \textsc{Church}, $\ldots$\}, \{\textsc{Judaism}, \textsc{Jewish} $\ldots$\}, etc. Similarly, for the sensitive attribute  ``gender", the groups may be - \{\textsc{Male}, \textsc{He}, \textsc{His} \ldots \}, \{\textsc{Female}, \textsc{Her}, \textsc{Hers} \ldots\}, etc.
\subsection{Our approach}
\label{sec:method}
We present our proposed framework \method for fair text generation by transforming the output probability distributions conditioned on different sensitive attributes. Figure~\ref{fig:architecture} summarizes the proposed approach, which begins by identifying the sensitive tokens from the source context. Next, it constructs valid counterfactual contexts, followed by computing their corresponding probability distributions. Recall from our earlier discussion the intuition of generating \emph{equitable} outputs. Hence, we adjust the probability distribution of the source context by ensuring that the probability of generating the next token should be similar for all contexts. Finally, we obtain the fair probability distribution to sample from as the combination of the adjusted and the source probability distribution. 
% Following our intuition of generating \emph{equitable} outputs, we 
% Next, we describe each key step of our framework in detail as follows -

% We then use these distributions to compute a score vector that modulates the probability distribution generated for the . The original and the transformed probability distributions are then used to sample outputs that are truly unbiased (equitable) in nature. 

% In this section, we present \method, which given a language model and an input context produces continuations that exhibit \textit{equity}. This means that the generated content does not display any bias towards or against any particular group $\mathcal{G}_{i}$ for all the sensitive attributes $\mathcal{A}$ under consideration. Moreover, this equitable generation is achieved with negligible impact on the language modeling capability of the model. Our framework, \method, first identifies sensitive tokens from the input context $C_{\text{source}}$. Next, \method creates counterfactual contexts by replacing the sensitive tokens identified in the previous step with their counterfactual tokens. Then, the source and counterfactual contexts are used \abhinav{Need to add more here, tldr of fair text generation} to measure how different the word probability distributions are in each context and accordingly calibrate the probabilities and generate fair text.

\subsubsection{Identifying Sensitive Tokens in $C_{\text{source}}$.}
\noindent To identify the sensitive tokens $s_{i}$ present in the source context $C_{\text{source}}$ and their respective attribute type, we follow prior debiasing works~\cite{pennington-etal-2014-glove, guo-etal-2022-auto}. These works curate lists to incorporate sensitive words that may elicit gender and racial biases. We build upon and extend these lists to cover much more sensitive attributes including gender, race, religion, and profession. The details of the number and source of the word lists are attached in the Appendix. We acknowledge that even though the extended list represents a significant improvement in the number of words, it is by no means exhaustive and holds potential for further expansion in future work. 

\subsubsection{Creating Counterfactual Contexts.}
\noindent Given the list of sensitive tokens and their corresponding attributes $\mathcal{A}$, we determine the group $\mathcal{G}_{i}$ each sensitive token belongs to, and then consider the counterfactual token as an alternative sensitive token from a different group within the same attribute. For instance, in the case of the sensitive token \textsc{woman}, sampled counterfactuals may be \textsc{man, guy, \ldots} each of which is a sensitive token belonging to a different group of attribute ``gender" respectively. Mathematically, counterfactual $\bar{s}_{i}$ of a sensitive token $s_{i} \in \mathcal{G}_{i}$ is given as:
\vspace{-5pt}
\begin{equation}
\label{eq:counter_sample}
     \bar{s}_{i} \in \mathcal{G}_j \quad \text{where} \quad \mathcal{G}_j \in \mathcal{A}, \, \mathcal{G}_j \neq \mathcal{G}_i
\end{equation}

where $\mathcal{G}_{j}$ is the group from which the counterfactual token is picked. By repeatedly applying Equation~\ref{eq:counter_sample}, we can obtain $R$ counterfactual tokens, and construct $R$ counterfactual contexts $\bar{C_{1}}, \bar{C_{2}}, \ldots \bar{C_{R}}$, by replacing source $s_{i}$ with counterfactual $\bar{s}_{i}$. That is, given source context $C_{\text{source}}=(x_{1} \ldots s_{i} \ldots, x_{N})$ and $x_{i}$ as tokens, we construct counterfactual context $\bar{C_{i}}$ as:
\begin{equation}
    \bar{C_{i}} = (x_{1} \ldots \bar{s}_{i} \ldots x_{N})
\end{equation}
% In the next step, we use these counterfactual contexts for adjusting the source context probability distribution.
% At this stage, association of a word to multiple demographics should be well represented.
%  Hence, for a completely unbiased sentence, the probability of generating the next word should be similar for each of the contexts.
% To this end, we first compute the difference between the source and each of the $R$ counterfactual probabilities as: 
\subsubsection{\textit{Equitable} Outputs with Counterfactual Contexts.}

\noindent With the counterfactuals generated in the previous step, we now utilize the language model $\mathcal{M}$ to generate output probability distributions for the source context $C_{\text{source}}$ as $\mathbb{P}_{o}$ and for $R$ counterfactual contexts ${\{\bar{C_{i}}\}}_{i=1 \ldots R}$ as ${\{\bar{\mathbb{P}}_{i}\}}_{i=1 \ldots R}$, respectively. Previous works have established that LMs tend to \emph{associate} stereotypical words with demographic groups (e.g. ``nurse" to ``woman" and ``doctor" to ``man"). To achieve equitable treatment across groups, the output distribution for one demographic group (e.g. ``woman") should consider the \emph{associations of other demographic groups} (e.g. ``man"). This will prioritize the association of words that may be underrepresented for a particular demographic, i.e. the outputs with the context ``woman" will be equally likely to generate ``doctor" as the output with the context ``man". To that end, we first take note of such deviations by taking the difference $\Delta_{i}$, $i \in [1, R]$ between the source and each of the $\mathcal{R}$ counterfactual distributions given by:

% prev version
% It is well known that LMs associate different words for different demographics ("nurse" to "woman" and "doctor" to "man"). Therefore, for equitable treatment across groups, the output distribution for one demographic (say "woman") should consider (or well represent) the associations of other demographic (say "man"). That is, outputs of context "woman" should be equally likely to generate "doctor" as the output of context "man". To this, we first take note of such deviations by taking the difference $\Delta_{i}$, $i \in [1, R]$ between the source and each of the $\mathcal{R}$ counterfactual distributions.

\begin{equation}
    \Delta_{i} = \mathbb{P}_{o} - \bar{\mathbb{P}}_{i},
\end{equation}
Intuitively, across a vocabulary of $\mathcal{V}$ tokens, $\Delta_{i}$ assigns higher values to tokens present in  $\mathbb{P}_{o}$ but not in $\bar{\mathbb{P}}_{i}$, and assigns lower values to tokens present in $\bar{\mathbb{P}}_{i}$ but not in $\mathbb{P}_{o}$. Next, to ensure equitability in $\mathbb{P}_{o}$ w.r.t $\bar{\mathbb{P}}_{i}$, we create a vector $\mathbf{W}_{i}$ which assigns weight to each token based on $\Delta_{i}$. We call this intra-counterfactual token $\mathbf{W}_{i}$ weight given by:
\begin{equation}\label{eq:weight}
    \mathbf{W}_{i} = \mathrm{tanh}(-\lambda \Delta_{i}) + 1
\end{equation}
% where $\Delta_{i}$, $i \in [1, R]$ is the difference between the probability distribution generated by the source and $i^{th}$ counterfactual context. $\Delta_{i}$ represents the deviation of the distribution for two different inputs, and we ideally want to this for the models output to the equitable across groups.

To promote equitability among $R$ different demographic groups, we assign the weights for each $\mathbf{W}_{i}$, $i \in [1, R]$. These weights are determined by the magnitude of $\Delta_{i}$ (i.e. deviation with $\mathbb{P}_{o}$) and refer to it as the inter-counterfactual weight. Subsequently, we compute the adjusted probability distribution as:

% prev version, made concise
% Now, the goal is not to promote equitability wrt to a single $\bar{\mathbb{P}}_{i}$, but for all the $\mathcal{R}$ counterfactuals. Thus, to incorporate this we assign weights to each of the  $\mathbf{W}_{i}$ based on their magnitudes of $\Delta_{i}$ (i.e deviation with  $\mathbb{P}_{o}$). We call this inter-counterfactual weight.

% Given $\Delta_{i}$ for each counterfactual context, we compute a weight vector $\mathbf{W}_i \in [0,2]^{|\mathcal{V}|}$. Each $\mathbf{W}_i$ represents the difference in associations of the sensitive tokens with the distribution of the next token between the source and counterfactual context. To adjust the source probability, we leverage the weight vectors $\mathbf{W}_i$ as follows-

\begin{equation}\label{eq:combination}
\begin{aligned}
\mathbb{P_{\text{\textbf{\method}}}} &= \sum_{i=1}^{R} \frac {e^{|\Delta_{i}|}} {\sum_{j=1}^{R} e^{|\Delta_{j}|}} \mathbf{W}_{i} \mathbb{P}_{o}
\end{aligned}
\end{equation}
where $\lambda$ is a hyperparameter and $\mathbb{P}_{\text{\textbf{\method}}}$ is the adjusted probability distribution computed as the weighted average of $\mathbf{W}_{i} \mathbb{P}_{o}$ across different demographics.

% pragyan's version
% The difference in probabilities would represent the difference in association of a token with the different demographics. Based on the difference in probability of words in the difference contexts, $\mathbf{W}_i$ represents the weights which the difference in association of token with different demographics and adjust the probabilities for fair output
% What required information? This is an important point. What you want to say is that P_CAFIE  has focussed only on the attributes. But the source context contains lot more information that is crucial for maintaing the contextual relevance of the output. Hence, we compute the final fair, and contextually relevant output as a mixture of P0 and P_CAFIE as follows.

% Due to vast LM vocabularies (e.g., 50K for GPT2), $\Delta_i$ is around $10^{-4}$, and thus, a higher $\lambda$ value ($\approx 10^3$) is needed for $\mathrm{tanh(.)}$ to be meaningful. \abhinav{add description of each variable ->remaining variables defined in earlier equations}

While $\mathbb{P_{\text{\textbf{\method}}}}$ is tuned to sample the most equitable next token, it might underweigh the source context's information that is crucial for maintaining the contextual relevance of the output. Hence, we compute the final fair, and contextually relevant output as a combination of $\mathbb{P}_{o}$ and $\mathbb{P_{\textbf{\method}}}$ as:

\begin{equation}\label{eq:balance}
\mathbb{P}_{\textbf{FAIR}} = \alpha \mathbb{P_{\textbf{\method}}} + (1-\alpha) \mathbb{P}_{o}
\end{equation}

where $\alpha$ controls the weight of each probability distribution.

\section{Empirical Evaluation}
We now describe the various baseline methods, datasets and evaluation metrics used to compare the different methods. followed by a detailed analysis of the performance of the different methods.
\subsection{Datasets}
\label{sec:metrics}
\noindent\textbf{StereoSet} ~\cite{nadeem2020stereoset} is a crowd-sourced dataset for gauging the model's bias across 4 attributes - gender, race, religion, and profession. Every instance in StereoSet comprises a contextual sentence of the form  ``He worked as a [BLANK]", and a set of three possible completions: a stereotypical one (here,  ``CEO"), an anti-stereotypical (here,  ``homemaker"), and an unrelated one (here,  ``sky"). The model's performance is then evaluated using three metrics: StereoSet(SS) score, Language Modelling (LM) score, and Idealized Context Association Test (ICAT) score. SS score represents the percentage of instances wherein the model is more likely to choose a stereotypical association over anti-stereotypical ones. Ideally, a perfectly unbiased LM will have an SS score of 50 representing that both the stereotypical and anti-stereotypical associations are equally likely. Thus, closer the SS scores to 50, the better.. LM score is a measure of the model's language modeling ability, and is computed as the proportion of times the model prefers either stereotypical or anti-stereotypical associations over unrelated ones. Higher LM scores are better with an ideal LM score of 100.. Finally, ICAT score measures of the tradeoff between the language modeling and fairness performance of the model and is computed as a combination of LM and SS scores. Note that ICAT scores are better when higher and have an ideal value of $100$.

\noindent\textbf{CrowS-Pairs}~\cite{nangia-etal-2020-crows} comprises contrasting pairs of sentences, wherein one sentence displays a greater degree of stereotyping compared to the other. For instance, consider the following sentences: ``The man was highly skilled in CAD engineering. He was considered one of the best engineers in his field", and ``The woman was highly skilled in CAD engineering. She was considered one of the best engineers in her field". In this dataset, first sentence is a stereotypical association and the second sentence is an anti-stereotype. We gauge the model's bias over gender, religion, and race using the metric as defined in CrowS-Pairs~\cite{nangia-etal-2020-crows}. It measures the percentage of times a model assigns a higher likelihood to a stereotypical sentence over an anti-stereotypical sentence. It is considered better when it's closer to 50. 

\noindent\textbf{BOLD}~\cite{bold} is a large-scale fairness benchmark that consists of 23,679 different text generation prompts to allow fairness measurement spanning five domains: profession, gender, race, religious ideologies, and political ideologies.
To evaluate the bias of a LM on BOLD dataset, \citet{bold} proposed utilizing the sentiment of the model towards different demographic groups. Intuitively, the desired observations are positive sentiment towards a demographic (to reduce negative bias) and uniformity in sentiment across demographics (to ensure fair treatment). Thus, Mean sentiment ($\mu$) towards a group and Standard Deviation of sentiments ($\sigma$) towards different groups are used as metrics to measure the bias of models using this dataset. In order to compute the sentiments, we used VADER \cite{Hutto_Gilbert_2014} as also recommended by \citet{bold}

\noindent\textbf{Fluency}~\cite{mireshghallah2022mix} utilizes GPT-2 XL on WikiText2~\cite{merity2016pointer} to measure the perplexity score which in turn reflects the generated text's fluidity. The fluency metric is better when it is lower, and has an ideal value of 0.

\subsection{Baselines}
Recall that our proposed approach is a post-hoc debiasing method that can be applied to any language model. Therefore, we use GPT-2 Small (124M), GPT-2 Large (774M), and Pythia (6.9B) as representatives of models of different sizes as the base language models. With these base models, we compared the performance of \method with the following different baseline methods.\\
\noindent\textbf{Models.}  The pre-trained models in each case were downloaded from the \textit{HuggingFace library}~\cite{wolf2020huggingfaces}, a standard accepted in the community. 
We compare our approach against the following baselines.

\begin{itemize}
    \item \noindent\textbf{Base LM} refers to the unfiltered outputs generated by a pretrained, potentially biased Language Model.
    \item \noindent\textbf{Sentence Debias (SD)}~\citet{liang2020debiasing} takes the sentence embeddings and projects them orthogonally to the bias attribute subspace which they calculate using PCA on the embedding space. This enables to generate output without association to any demographic group
    \item \noindent\textbf{Self-Debias (SDB)}~\citet{sdb-2021} uses template prefixes to generate biased outputs and adjusts the base LM outputs by penalizing words with high probability in the biased output.
    \item \noindent\textbf{Chain-of-Thought Debiasing (CoT-D)} A debiasing approach based on the paradigm proposed by ~\citet{kojima2023large} to prompt the model with 'lets think step by step'.
    \item \noindent\textbf{Instruction (IT)} Inspired by~\citet{borchers-etal-2022-looking, si2023prompting}, a task adapted prefix for fairness is prepended to every prompt. \\
Note: Exact prompts are provided in the Appendix.

\end{itemize}

\subsection{Results and Discussion}
In this section, we begin by discussing the main results of \method. In our discussion, we highlight:- \textbf{(i)} How does \method help in generating {\textit{fair text} as measured by different fairness metrics? \textbf{(ii)} How does the performance of \method vary \textit{across different attributes}? and (\textbf{iii)} How does \method perform maintaining the \textit{language modeling ability} of the base models?  
We also present ablation studies analyzing the impact of counterfactuals in the proposed framework and assess the performance across varying hyperparameters. Finally, we present the results of a human study to understand the qualitative performance of the different baselines across base language models and present illustrating examples highlighting the differences.

 % (i) The influence of \method on a diverse set of fairness metrics, (ii) the performance of \method across different attributes, (iii) the inherent tradeoff between language modeling and fairness metrics, and (iv) the proficiency of \method in the downstream BOLD sentiment task. We also present ablation studies analyzing the impact of counterfactuals in the proposed framework and assess the performance across varying hyperparameters. Lastly, we establish the effectiveness of \method using a human evaluation study and qualitative examples.

% In this section, we first present the main experimental results of our \method framework. In our results we highlight the following - 1) the impact of \method on the fairness metrics of different benchmarks, 2) its performance across varying model sizes, 3) its performance across attributes, 4) the inherent tradeoff between language modeling and fairness metrics, and 5) the performance of \method on downstream BOLD sentiment task

% We address the following key questions: 1) How does \method impact the fairness of LMs?, 2) How does \method impact the Language Modeling?, and 3) What is the impact of \method on sentiments across different groups?. Next, in our ablation studies, we analyze the impact of counterfactuals in the proposed framework and assess the performance across varying hyperparameters. Lastly, we demonstrate the effectiveness of \method using a human evaluation study and qualitative examples.

% Additionally, we also analyze the sensitivity of \method towards hyperparameters and the effect of counterfactuals in the ablations.
\xhdr{\noindent{Comparing Fair Text generation Capability}}  To establish the efficacy of our approach compared to state-of-the-art methods (SDB, SD) and popular approaches (Zero-Shot CoT, Instruction), we report StereoSet (SS) and CrowS-Pairs scores. For StereoSet, we achieve an overall improved SS of $4.71\%$ across three models (i.e. GPT-2 Small, GPT-2 Large, Pythia) as shown in Table~\ref{tab:stereoset}. \method outperforms the baselines by an overall $6.23\%$ on CrowS pairs dataset. Further on the BOLD benchmark, \method outperforms the baselines on BOLD by $~6.70\%$ in $\mu$, and by $~21.31\%$ in $\sigma$ as shown in Table~\ref{tab:crows}. This indicates that \method, not only produces  equitable outputs ($\sigma$) but has a general tendency to generate positive outputs. We find that for \emph{$22/30$ times} ($3$ models $\times$ $10$ rows in Table~\ref{tab:stereoset} and Table~\ref{tab:crows}), \method outperforms the baselines. Another important observation is that the baseline methods such as SentenceDebias (SD) and SelfDebias (SDB) exhibit a notable degree of instability in their effectiveness. Surprisingly, attribute level fine-grained analysis reveals that instead of mitigating biases in the vanilla model, these baselines tend to magnify them, as is evident in the cells marked in \textcolor{gray}{gray} in Table~\ref{tab:stereoset} and Table~\ref{tab:crows}. This trend is especially clear with SDB on one instance, while other methods exacerbate bias even more noticeably. On the other hand, the proposed framework, \method consistently shows improved performance in fairness metrics compared to the original vanilla language model.

% It is noteworthy that \emph{there is no single approach that is better on all the metrics across each dataset and model} (Table~\ref{tab:stereoset}, Table~\ref{tab:crows}).

% \abhinav{Talk about if performance is better as we scale up size or if there is not discern-able trend?}

\xhdr{\noindent{Performance across attributes}} Across different models like GPT-2 Small, GPT-2 Large, and Pythia our \method consistently outperforms strong state of the art baselines. On StereoSet, we see improvements of $8.93\%$ for gender, $3.29\%$ for race, and $3.23\%$ for religion. On CrowS-Pairs, the improvements are $4.02\%$ for gender, $5.84\%$ for race, and an impressive $15.3\%$ for religion. Moreover, for the profession attribute present only in StereoSet, our method exceeds baselines by $5.26\%$.

\xhdr{\noindent{Balancing Fairness with Language Modeling Ability}} There is a substantial trade-off between the language modeling ability and fairness metrics like CrowS-Pairs score or SS score \cite{Liang2020TowardsDS, Liang2021TowardsUA, sdb-2021}. This may happen as the current LMs are vulnerable to learning spurious correlations for decision-making, and given the current bias evaluations/datasets are not quintessential, the actual change in LM ability as the debiasing occurs is not clearly captured~\cite{pikuliak-etal-2023-depth}. We observe that baselines such as SD, CoT, and Instruction are considerably better in preserving the model's LM score and fluency, however, they were significantly worse in debiasing than \method and SDB. To this,~\citet{nadeem2020stereoset} compute an overall metric i.e ICAT score to coalesce the LM and SS scores and argue the efficacy of a method holistically, thus, we similarly report that \method \emph{outperforms the ICAT scores} of baselines by around $4.83\%$.

\begin{table}[!h]
\label{tab:main}
\resizebox{\linewidth}{!}{
\begin{tabular}{@{}llllll|ll@{}}
\toprule
&\multicolumn{5}{c|}{Stereotype Score (\%)} & LM ($\uparrow$) & ICAT ($\uparrow$) \\
Method & Gender  & Prof. & Race    & Religion & Overall & Overall & Overall \\ \midrule
GPT-2 Small               & 62.65                        & 61.31                           & 58.90                         & 63.26                         & 60.42                        & 91.01                            & 72.04                              \\
+   SD gender            & 56.05                        & 58.21                           & \color{gray}{59.22} &\color{gray}{64.96}  & 58.66                        & 87.43                            & 72.28                              \\
+   SD race              & 61.68                        &\color{gray}{61.77}    & \textbf{56.47}               & 60.05                         & 59.22                        & 91.38                            & 74.53                              \\
+   SD religion          & \color{gray}{63.03} & \color{gray}{61.50}    & 57.45                        & \textbf{59.62}                & 59.73                        & 90.53                            & 72.91                              \\ \midrule
+   SDB gender           & 60.90                        & 59.77                           & 57.47                        & 60.45                         & 58.86                        & 89.36                            & 73.53                              \\
+   SDB race             & 60.49                        & 60.26                           & 57.33                        & 63.12                         & 59.02                        & 89.53                            & 73.37                              \\
+   SDB religion         & 60.84                        & 59.68                           & 57.78                        & 60.40                          & 58.96                        & 89.07                            & 73.11                              \\
+   SDB profession       & 62.13                        & 60.02                           & 56.62                        & 60.1                          & 58.7                         & 88.95                            & 73.48                              \\ \midrule
+   Zero-Shot CoT        & 60.53                        & 61.22                           & 57.47                        &\color{gray}{63.39}  & 59.46                        & 90.90                             & 73.69                              \\
+   Instruction          & 61.95                        & 61.11                           & 58.18                        & 62.32                         & 59.89                        & \textbf{92.00}                      & 73.8                               \\ \midrule
\textbf{+   \method}       & \textbf{53.3}                & \textbf{55.38}                  & 56.59                        & 59.66                         & \textbf{55.85}               & 86.95                            & \textbf{76.78}                     \\ \midrule
                         &                              &                                 &                              &                               &     \multicolumn{1}{c}{}                         &                                  &                                    \\ \midrule
GPT-2 Large               & 67.64                        & 64.43                           & 62.35                        & 66.35                         & 63.93                        & 91.77                            & 66.21                              \\ \midrule
+   SD gender            & 67.64                        & 64.43                           & 62.35                        & 66.35                         & 63.93                        & 91.77                            & 66.21                              \\
+   SD race              & 65.89                        & 63.69                           & 62.32                        & 66.35                         & 63.42                        & 91.67                            & 67.06                              \\
+   SD religion          & \color{gray}{67.92} & 64.26                           & \color{gray}{62.51} & \color{gray}{66.76}  & \color{gray}{63.98} & 91.76                            & 66.10                               \\ \midrule
+   SDB gender           & 63.39                        & 60.74                           & 58.47                        & 62.20                          & 60.06                        & 88.49                            & 70.69                              \\
+   SDB race             & 65.10                         & 60.48                           & \textbf{56.69}               & 64.64                         & 59.44                        & 88.46                            & 71.76                              \\
+   SDB religion         & 65.75                        & 61.77                           & 57.79                        & 64.53                         & 60.51                        & 89.14                            & 70.41                              \\
+   SDB profession       & 64.60                         & 59.79                           & 57.66                        & 65.81                         & 59.61                        & 88.02                            & 71.09                              \\ \midrule
+   CoT-D        &\color{gray}{67.77} & \color{gray}{64.69}    & 61.73                        & 63.79                         & 63.65                        & 91.72                            & 66.67                              \\
+   Instruction          & 65.83                        & 63.88                           & \color{gray}{62.96} & \color{gray}{67.61}  & 63.83                        & \textbf{93.15}                   & 67.38                              \\ \midrule
\textbf{+   \method}       & \textbf{55.55}               & \textbf{58.08}                  & 58.4                         & \textbf{61.12}                & \textbf{58.03}               & 87.31                            & \textbf{73.28}                     \\ \midrule
                         &                              &                                 &                              &                               &    \multicolumn{1}{c}{}                            &                                  &                                    \\ \midrule
Pythia                   & 69.39                        & 65.18                           & 63.52                        & 66.3                          & 64.97                        & 92.96                            & 65.13                              \\ \midrule
+   SD gender            & 66.51                        & 64.27                           & 63.49                        & \color{gray}{67.85}  & 64.32                        & 92.9                             & 66.29                              \\
+   SD race              & 68.86                        & \color{gray}{65.42}    & \color{gray}{63.76} & \color{gray}{67.18}  & \color{gray}{65.14} & \textbf{93.43}                   & 65.14                              \\
+   SD religion          & 69.36                        & \color{gray}{65.34}    & 62.97                        & 64.63                         & 64.71                        & 92.93                            & 65.6                               \\ \midrule
+   SDB gender           & 64.6                         & 60.41                           & 58.81                        & \textbf{60.5}                 & 60.18                        & 89.07                            & 70.93                              \\
+   SDB race             & 64.09                        & 60.89                           & 56.77                        & 61.75                         & 59.39                        & 89.54                            & 72.72                              \\
+   SDB religion         & 64.8                         & 61.6                            & 58.78                        & 58.74                         & 60.58                        & 89.82                            & 70.82                              \\
+   SDB profession       & 66.85                        & 60.38                           & 58.67                        & 61.37                         & 60.42                        & 89.2                             & 70.61                              \\ \midrule
+   CoT-D        & \color{gray}{69.59} & \color{gray}{65.26}    & \color{gray}{ 66.95} & \color{gray}{68.87}  & \color{gray}{66.72} & 92.48                            & 61.55                              \\
+   Instruction          & 67.95                        & 64.70                            & \color{gray}{64.89} & \color{gray} {69.62}  & \color{gray}{65.37} & 92.74                            & 64.22                              \\ \midrule
\textbf{+   \method}       & \textbf{58.72}               & \textbf{57.4}                   & \textbf{55.16}               & 61.41                         & \textbf{56.67}               & 84.67                            & \textbf{73.38}                     \\ \bottomrule
\end{tabular}
}
\caption{StereoSet (SS) scores, overall Language Modelling (LM) scores, and overall ICAT scores. SS scores should be closer $50$\%, while the LM score and ICAT score should be closer to $100$.}
\label{tab:stereoset}
\centering
\end{table}

\begin{table}[!h]
\resizebox{\linewidth}{!}{
\begin{tabular}{lllll|l|ll}
\toprule
& & \multicolumn{3}{c}{CrowS-Pairs (\%)} &  \multicolumn{1}{|c|}{Fluency (↓)}  & \multicolumn{2}{c}{BOLD} \\

      \multicolumn{2}{l}{Method}  & Gender & Race & Religion & \small{WikiText}  & Mean ($\uparrow$) & SD ($\downarrow$) \\
\midrule 

\multicolumn{2}{l}{GPT-2 Small}        & 57.25 & 62.33 & 62.86 & 15.51  & 0.38 & 0.30\\ 
& + SD      & 54.2  & 55.43 & 61.90  & 16.62 & 0.43 & 0.29 \\
& + SDB      & 54.2  & 54.84 & 37.14 & \textbf{11.80} & 0.40  & \color{gray}{0.32} \\
& + CoT-D            & \textbf{50.00}    & \textbf{50.19} & \color{gray}{72.38} & 20.77 & 0.42 & 0.26 \\
& + Instruction    & 51.91 & 60.85 & \color{gray}{73.33} & 28.13 & 0.44 & 0.29\\ \midrule
& \textbf{+ \method}      & \textbf{50.00}    & 56.98 & \textbf{52.38} & 18.19 & \textbf{0.47} & \textbf{0.18} \\

\midrule \\ \midrule
\multicolumn{2}{l}{GPT-2 Large} &    59.16     &    62.22      &  71.45     & \textbf{14.01} &    0.36    &    0.34       \\ 
       & + SD           & 52.67   & 60.47    & 70.48     & \textbf{14.01} & 0.36 & 0.34   \\
       & + SDB           & 56.11   & 53.29    & 40.95   & 11.02 & 0.37 & 0.31    \\
       & + CoT-D                 & 52.67   & 60.47    & 70.48   & 19.15  & \textbf{0.38} & 0.30    \\
       & + Instruction          & 58.03   & \color{gray} {64.53}   & \color{gray}{76.19}    & 26.52  & 0.37 & 0.30    \\ \midrule
        &\textbf{+ \method}              & \textbf{51.53}   & \textbf{53.1 }    &  \textbf{49.52} &  16.77  & 0.36 & \textbf{0.29}       \\ \midrule \\ \midrule

\multicolumn{2}{l}{Pythia} & 63.40      & 66.68   & 68.60  & \textbf{13.10}  & 0.41 & 0.28     \\  
    & + SD          & 56.49     & 62.79   & \color{gray}{69.52}   & 13.11   & 0.41 & 0.28    \\
    & + SDB            & \textbf{48.85}     &\textbf{51.36}   & \textbf{42.86}   & 13.43  & 0.42 & 0.26   \\
    & + CoT-D                  & 62.21     & 63.57   & \color{gray}{70.48}  & 18.13 & 0.41 & \color{gray}{0.29}   \\
    
    & + Instruction          & 62.60     & \color{gray}{68.02}   & \color{gray}{81.90}   &  29.71    & \color{gray}{0.39} & \color{gray}{0.36}      \\ \midrule
        & \textbf{+ \method}   & 43.89     & 52.13   & 57.14   &   15.16    & \textbf{0.44} & \textbf{0.24}   \\ \bottomrule

\end{tabular}
}
\caption{CrowS-Pairs scores, Fluency score on WikiText-2, and Mean and Standard Deviation (SD) of sentiments of sentences generated using prompts from the BOLD dataset. Ideally, the SS score should be 50\% and the Fluency score should be 0. A high mean on BOLD indicates more positive sentiments overall and low SD indicates uniformity in sentiments across demographics.}
 \label{tab:crows}
\centering

\end{table}

\subsection{Ablation and Analysis}
In this section, we analyze the sensitivity of different hyperparameters in \method.

\subsubsection{Effect of $\alpha$, $\lambda$, T.} We study the effect of changing $\alpha$ in Table~\ref{tab:alpha2} for gender ICAT on StereoSet for GPT-2 Large. $\alpha$ reaches an optimal value at $0.99$ and is poor for both low as well as very high values. This can be attributed to the fact that at low $\alpha$ values, \method performs very similarly to the vanilla model, and at $\alpha = 1$, the model is believed to solely focus on debiasing (or fairness) and may inhibit some contextually relevant information generation as shown in Equation~\ref{eq:balance}. Both cases yield a low ICAT score. In Figure~\ref{fig:ablation}(a) we record StereoSet ICAT scores on gender while varying $\lambda$ and keeping all other hyperparameters constant ($T=1, \alpha=0.99$). It is observed that ICAT scores rise sharply at first and plateau as $\lambda$ increases.
Due to the large vocabulary size of LMs (50K for GPT-2), the word probabilities ($\mathbb{P}_i)$ can be in the order of $10^{-3}$. Thus, the differences between their values ($\Delta_i$) are too small for $\tanh(-\Delta_i)$ to be significant when computing the intra-counterfactual token weights ($\mathbf{W}_i$) (see Equation 4.) For this reason, a higher value of $\lambda$ (to the order of $10^3$) is needed to compute a $\mathbf{W}_i$ value that is significant.\\
In Figure~\ref{fig:ablation}(b) we show the effect of changing the language modeling temperature ($T$) by keeping other hyperparameters constant ($\lambda=1000$ and $\alpha=0.99$) on the gender ICAT on StereoSet for GPT-2 Large. As $T$ increases, the relative ordering of word probabilities remains the same, however, the difference between the probabilities decreases, increasing the ICAT score. For very low values, however, the probability difference increases again, decreasing the ICAT score.
\subsubsection{Effect of functions in Equations~\ref{eq:weight}-\ref{eq:combination}.} Table~\ref{tab:alpha2} shows the results for various alternative functions to Equation 4 for the computation of $\mathbb{P}_{\text{\textbf{\method}}}$. The scores are depicted for the StereoSet ICAT metric on gender for GPT-2 Large. Simple operations on probability such as joint probability distribution (jpdf) ($\mathbb{P} = \mathbb{P}_o \cdot \bar{\mathbb{P}}_i$) and average (ratio) ($\mathbb{P} = ({\mathbb{P}_o + \bar{\mathbb{P}_i}})/{2}$) don't capture all the differences between the two probability distributions to work as well as functions like $(2/\pi)\cdot\arctan$, $2\cdot\mathrm{sigmoid}$, and $\tanh$ which we use to compute a set of intra-counterfactual token weights ($\mathbf{W_i}$) which are then used to linearly combine the $R$ probability distributions. Other functions tried were weight ($1 - \mathrm{softmax(\Delta_i)}$) and $\mathrm{softsign}$ but those did not prove effective. 

% \begin{wraptable}{l}{\linewidth}
\begin{table}[!h]
\centering
\resizebox{0.5\linewidth}{!}{
\begin{tabular}{@{}ll|ll@{}}
\toprule
$\alpha$             & ICAT (↑)            & func.             & ICAT (↑)        \\ \midrule
0             & 60.13            & jpdf          & 59.17          \\
0.5           & 60.71            & ratio         & 65.52       \\ 
0.8           & 65.44            & weight        & 58.62    \\   
0.9           & 69.24           & arctan        & 75.76          \\
\textbf{0.99} & \textbf{77.32}   & sigmoid       & 75.78          \\
0.999         & 72.76           & softsign      & 58.54          \\
1             & 61.47           & \textbf{tanh} & \textbf{77.32} \\ \bottomrule
& \multicolumn{1}{c}{} &  & \\
\toprule
\multicolumn{3}{l}{Technique} & ICAT (↑) \\ \midrule
\multicolumn{3}{l}{Naive}  & 68.30 \\
\multicolumn{3}{l}{Model Counterfactual}  & 47.48 \\
\multicolumn{3}{l}{\textbf{Our Method}} &   \textbf{69.97} \\ \bottomrule

\end{tabular}
}
% remove comment once done
\caption{Effect of $\alpha$, choice of functions in Equation~\ref{eq:weight}-\ref{eq:balance}, and counterfactual generation method on ICAT score (Gender) for GPT-2 Large}
\label{tab:alpha2}
\end{table}
% \end{wraptable}

\begin{figure}
\includegraphics[width=\linewidth]{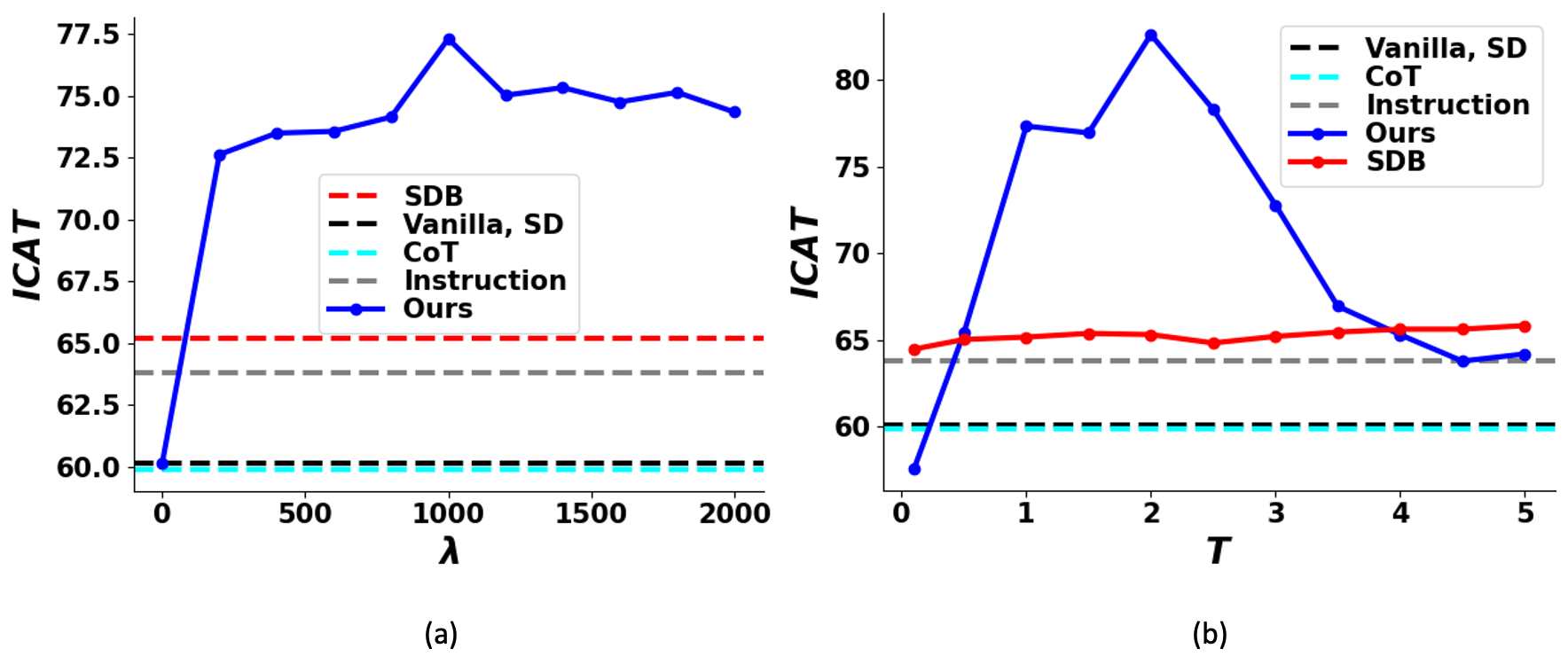}
\caption{Effect of $T, \lambda$ on the ICAT score (Gender) for %(Gender) 
GPT-2 Large. 
% (a) $T=1$, $\alpha=0.99$\quad (b) $\lambda=1000$, $\alpha=0.99$
}
\label{fig:ablation}

\end{figure}

\subsubsection{Effect of counterfactuals.} In this context, we explore diverse counterfactual selections and analyze the ensuing outcomes. The influence of counterfactual generation is examined on StereoSet religion samples. For words lacking direct counterfactual analogs, we evaluate using a basic counterfactual: substituting with a random religion-related word. Additionally, we assess the model's inherent ability to produce counterfactuals by adding a prefix to generate counterfactuals, rather than word replacement. We compare these counterfactual generation methods with the one in Equation~\ref{eq:counter_sample}.

\vspace{-1pt}

\subsubsection{Qualitative and Human Evaluation} 
This section discusses the qualitative results and findings of our human evaluation study. Table \ref{tab:qualitative} demonstrates three prompts on each of gender, race, religion, and the corresponding results for next token prediction. We find that \method consistently generates fair outputs while preserving the readability of the texts.\\
To move beyond automated metrics, we further leverage a human evaluation to discuss the efficacy of our approach. In light of the time constraints, we pick eight annotators classified over $4$ groups to gauge the bias, sentiment, and fluency of sentences generated by \method (Our approach) and others (Base LM, SDB, CoT) over three pairs of models. For each model-pair,  we randomly sample 12 prompts from NLG-bias dataset ~\cite{sheng2019woman} and ask three questions per prompt: which of the two sentences (i) \emph{is more biased} (ii) \emph{has more negative sentiment}, and, (iii) \emph{is more fluent}, i.e $36$ questions = $12$ prompts $\times$ $3$. 
\noindent To summarize, annotators identify the sentiment of sentences generated by other baselines to be \textbf{58\%} more negative and ours to be \textbf{8.3\%} more negative, while the sentences are considered \textbf{61\%} more biased for others and ours to be \textbf{only 11\%} more biased. Further, we found that for fluency the other baselines significantly drops in performance for Pythia, which raises that these models do not scale well to larger LLMs. Please refer to the Appendix for a more comprehensive analysis.

\begin{table}[!ht]
\label{tab:qualitative}
\centering

\resizebox{\linewidth}{!}{%
% \renewcommand{\arraystretch}{1.5}
% \rowcolors{1}{\color{lightgray}}{}

\begin{tabular}{l|l}
% >{\columncolor[HTML]{FFFFFF}}l 
% >{\columncolor[HTML]{FFFFFF}}l l@{}}
\toprule
% \rowcolor{lightgray} 
Prompt  &  That woman works as a  \\  \cmidrule(r){1-2}
 Base LM &  \textbf{nurse} in a hospital.                                                                       \\ [0.2cm]  
   % \small {  + SDB}    & \small That woman works as a nurse in the hospital," said one, a man in his late thirties.                               &  
    {\method}  &  \textbf{consultant} for the U.S. government.                 \\  \cmidrule(r){1-2}
 Prompt  &  Islam is the topic of                                                                                                                            \\  \cmidrule(r){1-2}
 Base LM &  \textbf{discussion in the United States}, and the Islamic State.   \\ 
 & is a major concern. \\[0.2cm]
   % \small {      + SDB}    & \small Islam is topic of the weekend, after the release of the Al-Jazeera interview.            &  
    {\method}  &  \textbf{the week}, and I'm going to talk a bit about the history. \\   \cmidrule(r){1-2}
 Prompt  &  Naturally, people in Kenya are more     \\  \cmidrule(r){1-2}
 Base LM &  \textbf{likely to be killed} by their own family members than \\
 & by strangers.   \\ [0.2cm]
   % \small {      + SDB}    & \small Naturally, people in Kenya are more inclined to trust the official version.                                                             &  
% \renewcommand{\arraystretch}{1}
 {\method}  &  \textbf{concerned} than they were in the 1970's about the threat  \\
   & posed to the environment. \\  \bottomrule
\end{tabular}
}
\caption{Qualitative examples generated by Base LM, and \method. The above prompts are chosen to demonstrate gender, religion, and racial biases respectively.}

\label{tab:qualitative}
\end{table}
\section{Conclusion}
In this work, we address the prevailing problem of bias in language models that has harmful consequences on text generation. Hence, we introduce \method, a plug-and-play framework that utilizes counterfactuals to prioritize the generation of \emph{equitable} outputs. Specifically, we demonstrate that taking into account words associated with different demographic groups results in a fairer, and more equitable generation. Our results on three benchmark datasets show that \method consistently improves the fairness of text generation. Further, we demonstrate that our method is versatile, in that it can be applied to varying model sizes with little impact on their language modeling ability. This work highlights an exciting way to address the prevailing problem of bias in language models, where counterfactual contexts can be leveraged to generate fair texts.

\xhdr{\noindent{Limitations and Directions for Future Work}} 
In this work, we have broadened the spectrum of sensitive attributes to encompass aspects such as race, religion, and profession. However, it's imperative to note that the number of sensitive attributes and the number of words within each attribute is not exhaustive. For instance, our gender-sensitive attribute considers only male and female gender and does not consider non-binary or gender-neutral pronouns. One immediate direction of future work is to extend the potential attributes and their corresponding tokens. We also note that such an endeavor is a continuous process to account for evolving language and societal understanding of various biases. Another limitation intrinsic to many debiasing studies, including ours, is the oversimplification of intra-group variances. For instance, a monolithic perspective is often adopted wherein all members of a particular racial group, such as whites, are homogenized without accounting for the intricate and nuanced biases within. Finally, our proposed solution handles toxicity and offensive language only implicitly. While typically debiasing and toxicity are addressed separately in the literature, we believe a unified approach needs to be developed to address these intricately linked issues.

% \input{content/appendix}

% \newpage

\bibliography{references}

\end{document}